\documentclass[runningheads]{llncs}
\usepackage{graphicx}
\usepackage{multirow}
\usepackage{booktabs}
\usepackage{xcolor}

\begin{document}
\title{Does pre-training on brain-related tasks results in better deep-learning-based brain age biomarkers?}
\titlerunning{Does brain-related pre-training results in better brain age biomarkers?}
%


\author{
Bruno M. Pacheco\inst{1} \and 
Victor H. R. de Oliveira\inst{1} \and
Augusto B. F. Antunes\inst{2} \and
Saulo D. S. Pedro\inst{3} \and
Danilo Silva\inst{1}, 
for the Alzheimer’s Disease Neuroimaging Initiative\thanks{
Data used in preparation of this article were obtained from the Alzheimer’s Disease Neuroimaging Initiative (ADNI) database (adni.loni.usc.edu). As such, the investigators within the ADNI contributed to the design and implementation of ADNI and/or provided data but did not participate in analysis or writing of this report. A complete listing of ADNI investigators can be found at:
\url{http://adni.loni.usc.edu/wp-content/uploads/how\_to\_apply/ADNI\_Acknowledgement\_List.pdf}
}}
\authorrunning{B. M. Pacheco et al.}
\institute{
Federal University of Santa Catarina (UFSC), Florianópolis, SC, Brazil \\ \email{bruno.m.pacheco@posgrad.ufsc.br,victoroliveira.eng@hotmail.com,\\ danilo.silva@ufsc.br} \and
Alliar - NEPIA, Belo Horizonte, MG, Brazil \\ \email{augusto.antunes@alliar.com} \and
3778 Healthcare, Belo Horizonte, MG, Brazil \\ \email{saulo.pedro@3778.care}
}

\maketitle              
\begin{abstract}
Brain age prediction using neuroimaging data has shown great potential as an indicator of overall brain health and successful aging, as well as a disease biomarker.
Deep learning models have been established as reliable and efficient brain age estimators, being trained to predict the chronological age of healthy subjects.
In this paper, we investigate the impact of a pre-training step on deep learning models for brain age prediction.
More precisely, instead of the common approach of pre-training on natural imaging classification, we propose pre-training the models on brain-related tasks, which led to state-of-the-art results in our experiments on ADNI data.
Furthermore, we validate the resulting brain age biomarker on images of patients with mild cognitive impairment and Alzheimer's disease.
Interestingly, our results indicate that better-performing deep learning models in terms of brain age prediction on healthy patients do not result in more reliable biomarkers.

\keywords{Brain Age \and Deep Learning \and ADNI \and BraTS \and MRI \and Transfer Learning.}
\end{abstract}
\section{Introduction}

As human lifespan increases, there is a growing need for reliable methods to assess brain health and age-related changes in the brain.
Brain age prediction is a promising technique that uses neuroimaging data to estimate the apparent age of an individual's brain, which can serve as an indicator of overall brain health and successful aging, as well as a disease biomarker~\cite{cole_brain_2018,jonsson_brain_2019,liem_predicting_2017,cole_prediction_2015,gaser_brainage_2013}.
Deep learning models have shown great potential in accurately predicting brain age from magnetic resonance imaging (MRI) data~\cite{cole_predicting_2017,bashyam_mri_2020,peng_accurate_2021,poloni_deep_2022,jonsson_brain_2019}.

Training deep learning models for brain age prediction shares several challenges with other neuroimaging tasks, in comparison to traditional computer vision, such as the increased GPU memory used from the 3D data and the extensive pre-processing required to account for the variability in the acquisition process.
In particular, available neuroimaging datasets are much smaller than existing natural imaging datasets~\cite{russakovsky_imagenet_2015,hutchison_what_2010}, and deep learning models are known to be very dependent on sample sizes.
Therefore, data-efficient training strategies are crucial to achieve high performance in brain age prediction.

In this paper, we explore the impact of pre-training deep learning models for brain age prediction.
Inspired by the learning process of expert neuroradiologists, we apply transfer learning by pre-training our brain age models on a brain-related task.
For comparison, we also train models without pre-training and models pre-trained on natural image classification.
We investigate the performance gain from pre-training and evaluate the models' brain age prediction as a biomarker for cognitive impairment.
More specifically:
\begin{itemize}
    \item We pre-train deep learning models on the brain tumor segmentation task and compare them to models without pre-training and with pre-training on the ImageNet natural image classification task;
    \item We test the brain age models using data from the ADNI studies, and show that the models pre-trained on the brain-related task outperform the other models, achieving the state-of-the-art in brain age prediction;
    \item We evaluate the brain age prediction of all models as a biomarker for different clinical groups (healthy, mild cognitive impairment, and Alzheimer's disease);
    \item Our experiments suggest that, despite the common practice, better models in terms of brain age prediction of healthy patients do not result in more reliable biomarkers;
    \item All of our results are reported on a standardized, publicly available dataset, providing an easy comparison with future research.
\end{itemize}

\section{Related Work}

Detecting aging features on brain MRI has been an active area of research for many years~\cite{franke_estimating_2010,kondo_age_2015,wang_age_2014}.
The use of deep learning for brain age prediction has gained considerable attention in recent years~\cite{poloni_deep_2022,lee_deep_2022,popescu_local_2021,peng_accurate_2021,dinsdale_learning_2021,armanious_age-net_2021}.
In this section, we provide a brief overview of related works on brain age prediction from MRIs using deep learning models.

One of the earliest applications of deep learning to brain age prediction was presented by Cole et al.\,\cite{cole_predicting_2017}.
The authors employed a neural network comprising a convolutional backbone and a fully connected regression head to analyze a dataset of T1-weighted MRI scans from 2001 healthy subjects aged 18 to 90.
The training is performed solely on images of healthy subjects, following the hypothesis that the brain age of healthy subjects is close to their actual age.
Their deep learning model outperformed the machine learning approach (Gaussian Process Regressor).
The authors also assessed the reliability of the predictions across individuals and acquisition methods.

Jonsson et al.\,\cite{jonsson_brain_2019} developed a deep learning model for brain age prediction using brain MRI scans from 1264 healthy subjects aged 18 to 75.
They explored the impact of training and testing on distinct datasets, finding that the performance of brain age prediction degraded when the target dataset differed from the training dataset.

Bashyam et al.\,\cite{bashyam_mri_2020} proposed to improve brain age prediction by utilizing a larger dataset of brain MRI scans from 14,468 subjects.
The dataset included data acquired from different sites following different protocols, with subjects aged 3 to 95.
The authors pre-trained their neural network on the ImageNet dataset, which is an even larger dataset of natural images.
They found that models performing well at chronological age prediction might not be the best at providing brain age estimates that correlate to the diagnosis of diseases such as schizophrenia and Alzheimer’s.

Peng et al.\,\cite{peng_accurate_2021} proposed quality brain age prediction using a lightweight deep learning model.
They used a dataset containing 14,503 subjects from the UK Biobank, with ages ranging from 44 to 80.
The authors showed that even though larger models perform well on natural image tasks, smaller models can perform equally well and sometimes even better on medical imaging tasks.

Multiple authors have reported brain age performance on MRI data from ADNI~\cite{lam_3d_2020,more_brain_2023,poloni_deep_2022,lee_deep_2022,ly_improving_2020}.
To the best of our knowledge, neither has provided means to reproduce the dataset used for testing the models.
The studies either used a random, non-disclosed split, or did not provide which images have been selected from the ADNI database.
Therefore, a direct comparison is not possible, as we cannot perfectly replicate the evaluation setting.
Nonetheless, we highlight that the best performance reported was a mean absolute error of 3.10 years~\cite{lee_deep_2022}.
Further details on the performance of each approach can be seen in Table \ref{tab:adni-sota}.

\begin{table}[h]
\caption{Performance of brain age prediction methods on MRIs from ADNI. The authors have not disclosed from which phases of the ADNI study the subjects were drawn. A$\beta$(-) indicates that the subjects have sustained a negative amyloid beta status over 3 years.}\label{tab:adni-sota}
\centering
\begin{tabular}{l l l}
\toprule
                         & Test set                                   & MAE  \\
\midrule
Lam et al.\,\cite{lam_3d_2020}       & 631 CN subjects (10-fold cross validation) & 3.96 \\
Ly et al.\,\cite{ly_improving_2020} & 51 CN subjects with A$\beta$(-)            & 3.7  \\
More et al.\,\cite{more_brain_2023}   & 209 CN subjects                            & 6.56 \\
Lee et al.\,\cite{lee_deep_2022}     & 330 CN subjects                            & 3.10 \\
Poloni et al.\,\cite{poloni_deep_2022}  & 151 CN subjects over 70 years old          & 3.66 \\
\bottomrule
\end{tabular}
\end{table}

Overall, these studies demonstrate the potential of pre-training for brain age prediction and the need for more efficient training strategies, as acquiring medical imaging data is very laborious\footnotemark.
\footnotetext{In contrast to natural imaging datasets, as medical imaging requires an expensive procedure and legal authorization from each subject.}
None of them, however, took advantage of models trained for other brain-related tasks, such as brain tumor segmentation.
Previous works also lack a standardized dataset on which we could perform a fair comparison, either because they use private datasets or because they do not share which samples were used for training or testing.
Therefore, our paper stands out by comparing brain age models pre-trained on brain tumor segmentation to models without pre-training or pre-trained on natural image classification. 
Furthermore, we experimented on a standardized and publicly available dataset, providing reproducible results.

\section{Materials and Methods}

\subsection{Data}\label{sec:data}

\subsubsection{ADNI}

The Alzheimer's Disease Neuroimaging Initiative (ADNI) was launched in 2003 with the primary goal of testing whether serial MRI, positron emission tomography (PET), other biological markers, and clinical and neuropsychological assessment can be combined to measure the progression of mild cognitive impairment (MCI) and early Alzheimer’s disease (AD)~\cite{petersen_alzheimers_2010}.
The ADNI database contains longitudinal data from clinical evaluations, cognitive tests, biological samples, and various types of imaging data, including MRI, functional MRI and PET.
For up-to-date information, see \url{adni.loni.usc.edu}.

The study's data has been collected over several phases: ADNI-1, ADNI-GO, ADNI-2, and ADNI-3.
The dataset contains cohorts of individuals with AD, MCI, and healthy controls (Cognitively Normal, CN).
Furthermore, all exams underwent quality control for the image quality (e.g., subject motion, anatomic coverage).
In this paper, we employed available MPRAGE T1-weighted MRIs from all phases, filtering out images deemed “unusable” by the quality control assessment.
The images are available after gradient non-linearity and intensity inhomogeneity correction, when necessary.
An overview of the demographics from each dataset can be seen in Table~\ref{tab:adni-meta}.
More detailed information on the images from ADNI-1 used in this work can be found in Wyman et al.\,\cite{wyman_standardization_2013}.

\begin{table}
\caption{Overview of the subjects from the ADNI database whose images were used in this work. Age is considered at the time of the first visit.}\label{tab:adni-meta}
\centering
\begin{tabular}{ l c c c c c }
\toprule
& \,CN\, & \,MCI\, & \,AD\, & \,Age Range (mean)\, & Sex \\
\midrule
ADNI-1 & 229 & 401 & 188 & 55-91 (75.28) & 342 F / 476 M \\
ADNI-GO & 0 & 142 & 0 & 55-88 (71.25) & \,\,\,67 F / \,\,\,75 M \\
ADNI-2 & 150 & 373 & 109 & 55-91 (72.61) & 324 F / 307 M \\
ADNI-3 & 456 & 354 & 64 & 51-97 (72.78) & 475 F / 399 M \\
\bottomrule
\end{tabular}
\end{table}

\subsubsection{BraTS}\label{sec:data-brats}

The brain tumor segmentation challenge (BraTS)~\cite{menze_multimodal_2015} provides a dataset of structural brain MRIs along with expert annotations of tumorous regions~\cite{bakas_advancing_2017}.
For each subject, four MRI scan modalities are available: T1-weighted, contrast-enhanced T1-weighted, T2-weighted, and T2-Flair.
All images are available after preprocessing (registration to a common atlas, interpolation to 1 mm\textsuperscript{3}, and skull-stripping)~\cite{bakas_identifying_2018}.
In the 2020 edition\footnotemark, the dataset contained 369 images from subjects aged 18 to 86 (avg. of 61.2).
\footnotetext{https://www.med.upenn.edu/cbica/brats2020/data.html}

\subsection{Preprocessing}

We applied minimal preprocessing to the ADNI images.
Our major goal was to ensure all images would have the same orientation and spatial resolution, and that they would present no skull or non-brain-tissue information, i.e., only brain voxels would be present in the image.
Only the brain information must be available in the image, otherwise, the deep learning models could learn to predict the age based on other structures.

We register all ADNI images to the MNI152 template, interpolate to 2 mm³ resolution, and apply skull-stripping using HD-BET~\cite{isensee_automated_2019}.
To feed the preprocessed 3D MRI scans to the 2D brain age models, each volume was sliced through the axial plane.
We discard the slices from the top 40\,mm and the bottom 35 mm of the scan to exclude slices with little to no brain information.
Therefore, we extract 40 images from each 3D MRI scan.

With respect to the images from the BraTS dataset, no further (see Sec. \ref{sec:data-brats}) preprocessing is performed.
We feed all slices of the 3D MRI scan to the deep learning models.

\subsection{Deep Learning Models}

\begin{figure}
\includegraphics[width=\textwidth]{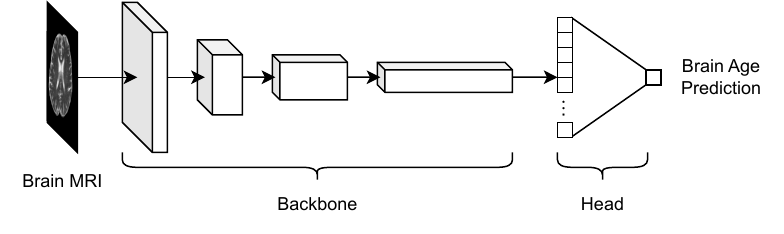}
\caption{Illustration of the overall architecture of our brain age model. The backbone is composed of several convolutional operations, extracted from the U-Net or the ResNet architectures. The head is a fully-connected layer applied to the vectorized values of the backbone's output.}\label{fig:model-diagram}
\end{figure}

We use 2D deep convolutional neural networks for brain age prediction.
Our proposed model can be divided into a backbone and a head.
Intuitively, the backbone is responsible for extracting relevant features from the input image, while the head combines these features into the final prediction.
The backbone consists of several convolution filters that reduce the image size.
The head is a single unit with linear activation that is fully connected to the backbone’s output.
Figure \ref{fig:model-diagram} illustrates our proposed model, highlighting both the backbone and the head.

Even though the architecture of the head is task-specific, the backbone’s architecture depends only on a few characteristics of the input (e.g., number of channels, minimum size).
Furthermore, learned features from one task can be useful for another, at least as a starting point, which is known as \textit{transfer learning}~\cite{raghu_transfusion_2019}.
This allows us to reutilize the backbone of models trained for different tasks, that is, we can extract the backbone from a model trained for some task and use it as the backbone for our model designed for brain age prediction.

We use the backbones from two different architectures: ResNet~\cite{he_deep_2015} and U-Net~\cite{navab_u-net_2015}.
More specifically, we use the ResNet-50 architecture, available in the torchvision package~\cite{marcel_torchvision_2010}, and the 2D U-Net proposed in \cite{isensee_nnu-net_2021}, which is designed for medical image segmentation.
The U-Net is composed of an encoder, a bottleneck, and a decoder.
For our backbone, we use the encoder with the bottleneck.

\begin{figure}[h]
    \centering
    \includegraphics[width=\textwidth]{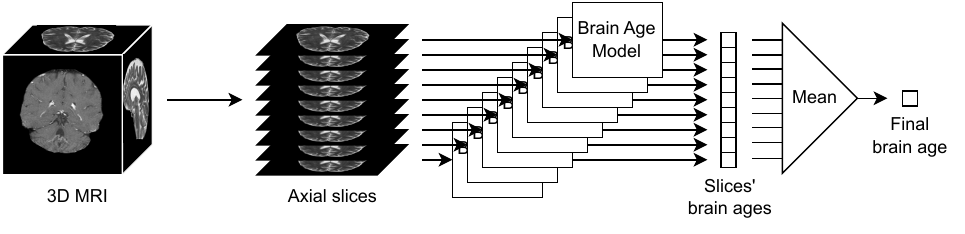}
    \caption{Brain age prediction of a 3D MRI using a 2D deep learning model.}
    \label{fig:pipeline}
\end{figure}

To obtain the brain age prediction of a 3D MRI, we apply the model to the 40 axial slices of the image that contain brain information and take the mean of the outputs~\cite{bashyam_mri_2020}.
The pipeline of operations can be seen in Figure \ref{fig:pipeline}

\subsection{Pre-training on brain tumor segmentation}\label{sec:brats-pretrain}

To leverage the knowledge from other brain-related tasks to brain age prediction, we pre-train our backbones in the brain tumor segmentation task.
We follow the BraTS challenge setup, with BraTS data, for both U-Net and ResNet backbones.
To be able to train a ResNet backbone in a segmentation task, we replace the original head of the ResNet with a U-Net decoder, in a ResUNet architecture~\cite{zhang_road_2018}.
This means that the decoder matches the ResNet backbone with respect to the size of the intermediate feature maps so that the skip connections can be added in the same way as in the original U-Net implementation.

We follow Crimi et al.\,\cite{crimi_nnu-net_2021} for training all models on the BraTS data.
We first train the models on a random 80/20 split of the BraTS 2020 dataset.
The models are evaluated through the Dice score~\cite{menze_multimodal_2015}, which measures the overlap between the predicted segmentation mask $\hat{Y}$ and the ground truth $Y$ as \[
    \textrm{Dice}(\hat{Y},Y) = \frac{2 |\hat{Y} \cap Y|}{|\hat{Y}| + |Y|}
.\]
Based on the models' performance, we fixed the number of epochs to avoid overfitting.
Then, the entirety of the BraTS data is used to train the backbones for a fixed number of epochs.

\subsection{Training on brain age prediction}\label{sec:brain-age-training}

To train the deep learning models on brain age prediction, we assume, following previous work, that the brain age of healthy subjects is close to their chronological age~\cite{cole_predicting_2017,bashyam_mri_2020,lee_deep_2022}.
Thus, we train and evaluate our models solely on the images of subjects belonging to the CN group.
To provide easy-to-compare results, we choose to evaluate our models on a standardized dataset.
Therefore, we follow the standardized split of the analysis set for ADNI-1~\cite{wyman_standardization_2013}, and use the standard test set as our test set, and their training set as our validation set.
In other words, we divide the preprocessed ADNI-1 T1-weighted scans from CN subjects following the standardized split to form our validation and test sets.
The remaining images (i.e., those from ADNI-GO, ADNI-2, and ADNI-3) compose our training set.
Detailed information on the images used in the training set can be found in our code repository\footnote{\url{https://github.com/gama-ufsc/brain-age}}.

We train all models, regardless of backbone architecture or pre-training, using the Adam optimizer to minimize the mean squared error between the age predicted from each slice and the true age of the CN subjects.
The models are first trained on the training set.
The performance of these models on the validation set is used for hyperparameter tuning and early stopping.
Namely, batch size and learning rate were adjusted, and a moving average\footnote{At the end of each epoch, we compute the average over the 5 latest results, including the current one.} of the MAE on the validation set was used to determine the ideal epoch (i.e., the one with the smallest MAE) for stopping the training.
The models with the best performance on the validation set are then evaluated on the test set, which is unseen up to then.

\subsection{Evaluation}\label{sec:evaluation}

The brain age models were evaluated on the error between the predicted age and the actual age of the CN subjects.
More specifically, we use the mean absolute error (MAE) as our standard evaluation metric.
We compute the error based on the predicted age of the whole 3D MRI, i.e., after averaging the predictions of all slices as described in Sec. \ref{sec:brain-age-training} and illustrated in Fig. \ref{fig:pipeline}.

Furthermore, we evaluate the capacity of the brain age estimate in differentiating between CN, MCI, and AD patients.
For this, we use the brain age delta $\Delta_{BA} = \hat{y}_{BA} - y_{CA}$, which is the difference between the predicted brain age $\hat{y}_{BA}$ and the chronological age $y_{CA}$ of a subject.
As the progression toward Alzheimer’s diagnosis is associated with aging patterns, it is expected that the $\Delta_{BA}$ of a subject in the AD group is greater than that of a subject in the MCI group, and that the latter’s $\Delta_{BA}$ is still greater than the $\Delta_{BA}$ of a subject in the CN group.
Therefore, we compute the predicted $\Delta_{BA}$ for all images in the three groups (CN, MCI, AD) of the test set and apply a pairwise Mann-Whitney U (MWU) test~\cite{mann_whitney_1947,michael_wilcoxon_2010}.
The MWU test is a nonparametric version of the t-test for independent samples.
In our case, the null hypothesis is that the $\Delta_{BA}$ from one group is not stochastically greater than the other.

\section{Experiments and Results}

In our experiments, we evaluate the impact of brain-related pre-training using 3 backbones: the U-Net with random initialization, ResUNet with random initialization, and ResUNet pre-trained on the ImageNet.
To improve the reliability of our results, 5 models are trained for each experiment, e.g., 5 U-Net models with random weight initializations are evaluated without brain-related pre-training, against 5 (different) U-Net models with brain-related pre-training.
In other words, for each of the 3 backbones, we train 10 models: 5 with no brain-related pre-training, and 5 with brain-related pre-training.


All experiments reported below were performed on a Linux machine with 8 vCPUs, 30\,GB of RAM and an Nvidia T4 GPU.
Further details regarding the implementation of the experiments and additional results can be seen in code our repository\footnote{\url{https://github.com/gama-ufsc/brain-age}}.

\subsection{Pre-training on BraTS}\label{sec:exp-brats}

Following the procedure described in Sec. \ref{sec:brats-pretrain}, we trained 5 U-Net models and 5 ResUNet models with random initialization on the brain tumor segmentation task.
We also used the backbone from ResNets pre-trained on the ImageNet’s natural image classification task, therefore, we trained 5 ResUNet models using the ImageNet pre-trained backbone.
ImageNet pre-trained models are readily available in the torchvision package.

Using an 80-20 random split, we observed that 30 epochs were enough to achieve peak performance and avoid overfitting for the U-Net models, while 50 epochs were enough for the ResUNet models.
The average performance of these models on the random split can be seen in Table \ref{tab:brats-score}.
We highlight that the models present performance on par with state-of-the-art brain tumor segmentation models\,\cite{bakas_identifying_2018}.
All models were then re-trained (with new initial random weights) on the entirety of the dataset for the same number of epochs.

\begin{table}[h]
\caption{Performance of the models on an 80-20 split of the BraTS dataset. Models were trained on 80\% of the data and evaluated on 20\%. Reported values are the average (and standard deviation) of 5 runs with random initialization, except for the backbone of the ResUNet that uses the backbone of the ResNet pre-trained on ImageNet on all runs.}\label{tab:brats-score}
\centering
\begin{tabular}{l c}
\toprule
Model & Dice Score \\
\midrule
U-Net & 0.8290 ($\sigma=0.0049$) \\ \hline
ResUNet & 0.8112 ($\sigma=0.0246$) \\ \hline
ResUNet (ImageNet) & 0.8129 ($\sigma=0.0181$) \\
\bottomrule
\end{tabular}
\end{table}

\subsection{Brain Age Prediction}\label{sec:exp-brain-age}

We train 5 models of each combination of backbone and pre-training available.
Namely: U-Net backbone with random initialization or pre-trained on BraTS; ResNet backbone with random initialization or pre-trained on BraTS; and ResNet backbone pre-trained on ImageNet or pre-trained on ImageNet and then on BraTS.
As described in Sec. \ref{sec:brain-age-training}, we use the validation set to define the best set of hyperparameters for each backbone and pre-training combination.
More specifically, we used a batch size of 64 images for all models and a learning rate of $10^{-3}$ for the models with U-Net backbone without pre-training, $10^{-5}$ for the models with ResNet backbone with ImageNet and BraTS pre-training, and $10^{-4}$ for all other models.
We trained all models for 50 epochs, early-stopping the training when a running average of the MAE on the validation set achieved the smallest value.
The average performance of the models on the test set can be seen in Table \ref{tab:brain-age-scores}.

\begin{table}[h]
\caption{Average performance (over 5 runs) of brain age models. “train+validation models” indicates that the models were trained on the union of train and validation datasets, with hyperparameters defined on previous experiments using the validation set. Values in bold indicate the best performance in their respective set (column).}\label{tab:brain-age-scores}
\centering
\begin{tabular}{l c c c c}
\toprule
\multirow{3}{*}{Backbone}                         & \multirow{3}{1.7cm}{\centering BraTS pre-training}   & \multirow{3}{*}{Validation MAE}           & \multirow{3}{*}{Test MAE}                 & \multirow{3}{3cm}{\centering Test MAE (train+validation models)} \\
& & & & \\
& & & & \\
\midrule
\multirow{2}{1.7cm}{U-Net}                             &                      & 3.203 ($\sigma=0.042$)          & 3.358 ($\sigma=0.099$)          & 3.138 ($\sigma=0.069$)                    \\ \cline{2-5}
                                                       & X                    & \textbf{3.186} ($\sigma=0.084$) & \textbf{3.284} ($\sigma=0.071$) & \textbf{3.079} ($\sigma=0.077$)           \\ \hline
\multirow{2}{1.7cm}{ResNet-50}                         &                      & 3.474 ($\sigma=0.117$)          & 3.556 ($\sigma=0.258$)          & 3.226 ($\sigma=0.071$)                    \\ \cline{2-5}
                                                       & X                    & 3.361 ($\sigma=0.079$)          & 3.413 ($\sigma=0.123$)          & 3.141 ($\sigma=0.030$)                    \\ \hline
\multirow{2}{1.7cm}{ResNet-50 (ImageNet)}              &                      & 3.509 ($\sigma=0.086$)          & 3.638 ($\sigma=0.160$)          & 3.210 ($\sigma=0.047$)                    \\ \cline{2-5}
                                                       & X                    & 3.452 ($\sigma=0.086$)          & 3.411 ($\sigma=0.066$)          & 3.238 ($\sigma=0.022$)                   \\
\bottomrule
\end{tabular}
\end{table}


Furthermore, after hyperparameter tuning, we re-train the models on the union of the training and the validation sets.
This increases the amount of data used for training and increases the similarity between the distribution of the training data and the test data, as both validation and test sets are drawn from the same study (ADNI-1).
We defined the training budget for each backbone and pre-training combination as the average of the epochs in which the respective 5 models achieved the early-stopping criterion on the validation set.
Note that the test set is not considered in any step of this process, therefore, no data leakage occurs.
The average test set performance of the models trained on the training and validation sets can be seen in Table \ref{tab:brain-age-scores}, under column ``Test MAE (train+validation models)''.

In our experiments, pre-training on the brain tumor segmentation task was consistently advantageous.
Even though the performance difference was not highly significant at all experiments, the models with the proposed pre-training outperformed their counterpart in all configurations except for the ResNet backbone pre-trained on the ImageNet when trained on the union of the train and validation sets.
It is also evident, when comparing the ResNet backbones, that pre-training solely on BraTS was consistently better than pre-training on the ImageNet.
Our experiments show that the best brain age predictions are achieved by using a U-Net backbone pre-trained on BraTS, even though the ResNet backbones achieved a better brain tumor segmentation performance (see Table \ref{tab:brats-score}).
Finally, the use of the validation set for training (after hyperparameter tuning) improved the performance of all models, as can be seen in the column ``Test MAE (train+validation models)'' of Table \ref{tab:brain-age-scores}.

\subsection{Statistical Analysis of $\Delta_{BA}$}\label{sec:exp-stats}

To assess the significance of the resulting brain age indicator, we apply the MWU test to the predicted $\Delta_{BA}$ of all models for subjects in the CN, MCI, and AD groups of the validation and the test sets (see Sec. \ref{sec:evaluation}).
The tests on samples between CN and MCI patients and CN and AD patients all pointed to a strong differentiation, with p-values smaller than 0.1\% for all models\footnote{The results of the tests on CN-MCI and CN-AD are available in our code repository \url{https://github.com/gama-ufsc/brain-age}.}, indicating that the $\Delta_{BA}$ biomarker is useful to distinguish healthy patients from those with Alzheimer's disease or mild cognitive impairment.
The distinction, however, between MCI and AD patients was not as significant, as can be seen in Table \ref{tab:brain-age-pvalues}.
Note that the images from MCI and AD patients of the validation set are not used for training or hyperparameter tuning, thus, the validation set 
can be interpreted as an additional
test set in the case of differentiating between MCI and AD.

\begin{table}[h]
\caption{Significance of the MWU test on the distinction between AD and MCI patients using the predicted $\Delta_{BA}$ values. We report the average (maximum) p-value over the 5 models trained for each combination of backbone and pre-training. “train+validation models” indicates that the models were trained on the union of train and validation datasets (CN only), with hyperparameters defined on previous experiments using the validation set (CN only). Values in bold indicate high significance (p-value $\le 5\%$).}\label{tab:brain-age-pvalues}
\centering
\begin{tabular}{l c c c c}
\toprule
Backbone                         & \multirow{2}{1.6cm}{\centering BraTS pre-training}   & Validation p-value           & Test p-value                 & \multirow{2}{4cm}{\centering Test p-value (train+validation models)} \\
& & & & \\
\midrule
\multirow{2}{1.6cm}{U-Net}                             &                      & \textbf{0.011} (\textbf{0.017}) & \textbf{0.025} (0.053)          & 0.091 (0.107)                    \\ \cline{2-5}
                                                       & X                    & \textbf{0.011} (\textbf{0.035}) & \textbf{0.036} (\textbf{0.049}) & 0.190 (0.213)                    \\ \hline
\multirow{2}{1.6cm}{ResNet-50}                         &                      & 0.051          (0.066)          & \textbf{0.046} (0.071)          & 0.192 (0.245)                    \\ \cline{2-5}
                                                       & X                    & 0.051          (0.064)          & 0.051 (0.089)                   & 0.261 (0.298)                    \\ \hline
\multirow{2}{1.6cm}{ResNet-50 (ImageNet)}              &                      & \textbf{0.044} (0.055)          & 0.063 (0.112)                   & 0.177 (0.230)                    \\ \cline{2-5}
                                                       & X                    & 0.095          (0.110)          & 0.171 (0.225)                   & 0.345 (0.390)                    \\
\bottomrule
\end{tabular}
\end{table}

Even though the models with the U-Net backbones achieved more significant results in the statistical analysis of the biomarker, these results do not allow us to conclude that pre-training (with any of the tasks) had a positive impact, as was observed for brain age prediction.
In fact, the exact opposite is observed.
Using pre-trained backbones (for both BraTS and ImageNet pre-training) resulted, most of the times, in models that achieved a worse separation between AD and MCI patients, in comparison to their counterparts without pre-training.
The same effect was also observed in the use of the validation set for training, upon which no model achieved a significant distinction between AD and MCI patients.

\section{Discussion and Conclusions}

In this study, we investigated the transfer learning capacity of a brain-related task to the task of brain age prediction.
More specifically, we pre-trained deep learning models for brain age prediction on the task of brain tumor segmentation.
In comparison to pre-training on a natural image classification task or performing no pre-training at all, our results 
suggest that the proposed pre-training may be a better option. 
The only inconclusive case is on performing both pre-trainings, that is, first pre-training on ImageNet and then on brain tumor segmentation, which yielded mixed results for brain age prediction.

Furthermore, we observed that using the validation set for training the models (after hyperparameter tuning) significantly reduced the brain age prediction error in all scenarios, even though there was only a small increment in the number of subjects in the training set (15\% more subjects, as can be verified through Table \ref{tab:adni-meta}).
Using the U-Net backbone pre-trained on BraTS and then trained on the union of train and validation sets showed state-of-the-art results, with MAE values previously unseen on ADNI data (see Table \ref{tab:adni-sota}).
We recall that the validation and test sets are built with images from the ADNI-1 study, while the training set is built from ADNI-GO, ADNI-2, and ADNI-3.
As the image acquisition protocols change between the studies, we can assume that there are different characteristics between their images.
Therefore, we can expect a distribution shift between the training and the test sets that does not exist between the validation and the test sets.
This allows us to conclude that the use of the validation set for training resulted in better models because it decreased the difference between the training and test distributions.

At the same time, by evaluating our models as biomarkers for cognitive impairment levels, we observed results that challenge the standard approach of training deep learning models for brain age.
Most of the modifications that improved the brain age predictive performance on healthy subjects (i.e., the chronological age prediction), resulted in models that were less reliable for distinguishing between patients with MCI and AD.
A similar behavior was reported by Bashyam et al.\,\cite{bashyam_mri_2020}, in which the models that achieved the lowest MAE did not provide the strongest distinction between healthy subjects and subjects with AD, MCI, Schizophrenia or Depression.

Our results show an inverse relationship between the performance on chronological age prediction (see Table \ref{tab:brain-age-scores}) and the reliability of the biomarker for cognitive impairment levels (see Table \ref{tab:brain-age-pvalues}).
This is particularly true for the use of the validation set for training brain age models, which significantly reduced the reliability in all scenarios.
Given that reducing the distribution shift between training and testing data degraded the performance of the biomarker, we speculate that state-of-the-art models got to a point of overfitting the images of healthy subjects, thus, degrading their performance on images from AD and MCI patients.
Therefore, we suggest that the validity of chronological age predictions as a means to develop brain age models is an important investigation topic for the development of the brain age biomarker.

\bibliographystyle{splncs04}
\bibliography{references}

\begin{thebibliography}{10}
\providecommand{\url}[1]{\texttt{#1}}
\providecommand{\urlprefix}{URL }
\providecommand{\doi}[1]{https://doi.org/#1}

\bibitem{armanious_age-net_2021}
Armanious, K., Abdulatif, S., Shi, W., Salian, S., Küstner, T., Weiskopf, D.,
  Hepp, T., Gatidis, S., Yang, B.: Age-{Net}: {An} {MRI}-{Based} {Iterative}
  {Framework} for {Brain} {Biological} {Age} {Estimation}. IEEE Transactions on
  Medical Imaging  \textbf{40}(7),  1778--1791 (Jul 2021).
  \doi{10.1109/TMI.2021.3066857}, conference Name: IEEE Transactions on Medical
  Imaging

\bibitem{bakas_advancing_2017}
Bakas, S., Akbari, H., Sotiras, A., Bilello, M., Rozycki, M., Kirby, J.S.,
  Freymann, J.B., Farahani, K., Davatzikos, C.: Advancing {The} {Cancer}
  {Genome} {Atlas} glioma {MRI} collections with expert segmentation labels and
  radiomic features. Scientific Data  \textbf{4}(1),  170117 (Sep 2017).
  \doi{10.1038/sdata.2017.117}

\bibitem{bakas_identifying_2018}
Bakas, S., Reyes, M., Jakab, A., Bauer, S., Rempfler, M., Crimi, A., Shinohara,
  R.T., Berger, C., Ha, S.M., Rozycki, M., et~al.: Identifying the {Best}
  {Machine} {Learning} {Algorithms} for {Brain} {Tumor} {Segmentation},
  {Progression} {Assessment}, and {Overall} {Survival} {Prediction} in the
  {BRATS} {Challenge}  (2018). \doi{10.48550/ARXIV.1811.02629}, publisher:
  arXiv Version Number: 3

\bibitem{bashyam_mri_2020}
Bashyam, V.M., Erus, G., Doshi, J., Habes, M., Nasrallah, I.M., Truelove-Hill,
  M., Srinivasan, D., Mamourian, L., Pomponio, R., Fan, Y., et~al.: {MRI}
  signatures of brain age and disease over the lifespan based on a deep brain
  network and 14,468 individuals worldwide. Brain  \textbf{143}(7),  2312--2324
  (Jul 2020). \doi{10.1093/brain/awaa160}

\bibitem{cole_brain_2018}
Cole, J.H., Ritchie, S.J., Bastin, M.E., Valdés~Hernández, M.C.,
  Muñoz~Maniega, S., Royle, N., Corley, J., Pattie, A., Harris, S.E., Zhang,
  Q., et~al.: Brain age predicts mortality. Molecular Psychiatry
  \textbf{23}(5),  1385--1392 (May 2018). \doi{10.1038/mp.2017.62}, number: 5
  Publisher: Nature Publishing Group

\bibitem{cole_prediction_2015}
Cole, J.H., Leech, R., Sharp, D.J., {for the Alzheimer's Disease Neuroimaging
  Initiative}: Prediction of brain age suggests accelerated atrophy after
  traumatic brain injury. Annals of Neurology  \textbf{77}(4),  571--581 (Apr
  2015). \doi{10.1002/ana.24367}

\bibitem{cole_predicting_2017}
Cole, J.H., Poudel, R.P.K., Tsagkrasoulis, D., Caan, M.W.A., Steves, C.,
  Spector, T.D., Montana, G.: Predicting brain age with deep learning from raw
  imaging data results in a reliable and heritable biomarker. NeuroImage
  \textbf{163},  115--124 (Dec 2017). \doi{10.1016/j.neuroimage.2017.07.059}

\bibitem{dinsdale_learning_2021}
Dinsdale, N.K., Bluemke, E., Smith, S.M., Arya, Z., Vidaurre, D., Jenkinson,
  M., Namburete, A.I.L.: Learning patterns of the ageing brain in {MRI} using
  deep convolutional networks. NeuroImage  \textbf{224},  117401 (Jan 2021).
  \doi{10.1016/j.neuroimage.2020.117401}

\bibitem{michael_wilcoxon_2010}
Fay, M.P., Proschan, M.A.: {Wilcoxon-Mann-Whitney or t-test? On assumptions for
  hypothesis tests and multiple interpretations of decision rules}. Statistics
  Surveys  \textbf{4}(none),  1 -- 39 (2010). \doi{10.1214/09-SS051}

\bibitem{franke_estimating_2010}
Franke, K., Ziegler, G., Klöppel, S., Gaser, C.: Estimating the age of healthy
  subjects from {T1}-weighted {MRI} scans using kernel methods: {Exploring} the
  influence of various parameters. NeuroImage  \textbf{50}(3),  883--892 (Apr
  2010). \doi{10.1016/j.neuroimage.2010.01.005}

\bibitem{gaser_brainage_2013}
Gaser, C., Franke, K., Klöppel, S., Koutsouleris, N., Sauer, H., {Alzheimer's
  Disease Neuroimaging Initiative}: {BrainAGE} in {Mild} {Cognitive} {Impaired}
  {Patients}: {Predicting} the {Conversion} to {Alzheimer}’s {Disease}. PLoS
  ONE  \textbf{8}(6),  e67346 (Jun 2013). \doi{10.1371/journal.pone.0067346}

\bibitem{he_deep_2015}
He, K., Zhang, X., Ren, S., Sun, J.: Deep residual learning for image
  recognition. In: 2016 IEEE Conference on Computer Vision and Pattern
  Recognition (CVPR). pp. 770--778 (2016). \doi{10.1109/CVPR.2016.90}

\bibitem{hutchison_what_2010}
Hutchison, D., Kanade, T., Kittler, J., Kleinberg, J.M., Mattern, F., Mitchell,
  J.C., Naor, M., Nierstrasz, O., Pandu~Rangan, C., Steffen, B., et~al.: What
  {Does} {Classifying} {More} {Than} 10,000 {Image} {Categories} {Tell} {Us}?
  In: Daniilidis, K., Maragos, P., Paragios, N. (eds.) Computer {Vision} –
  {ECCV} 2010, vol.~6315, pp. 71--84. Springer Berlin Heidelberg, Berlin,
  Heidelberg (2010). \doi{10.1007/978-3-642-15555-0\_6}, series Title: Lecture
  Notes in Computer Science

\bibitem{isensee_nnu-net_2021}
Isensee, F., Jaeger, P.F., Kohl, S.A.A., Petersen, J., Maier-Hein, K.H.:
  {nnU}-{Net}: a self-configuring method for deep learning-based biomedical
  image segmentation. Nature Methods  \textbf{18}(2),  203--211 (Feb 2021).
  \doi{10.1038/s41592-020-01008-z}

\bibitem{crimi_nnu-net_2021}
Isensee, F., Jäger, P.F., Full, P.M., Vollmuth, P., Maier-Hein, K.H.:
  {nnU}-{Net} for {Brain} {Tumor} {Segmentation}. In: Crimi, A., Bakas, S.
  (eds.) Brainlesion: {Glioma}, {Multiple} {Sclerosis}, {Stroke} and
  {Traumatic} {Brain} {Injuries}, vol. 12659, pp. 118--132. Springer
  International Publishing, Cham (2021). \doi{10.1007/978-3-030-72087-2\_11},
  series Title: Lecture Notes in Computer Science

\bibitem{isensee_automated_2019}
Isensee, F., Schell, M., Pflueger, I., Brugnara, G., Bonekamp, D., Neuberger,
  U., Wick, A., Schlemmer, H., Heiland, S., Wick, W., et~al.: Automated brain
  extraction of multisequence {MRI} using artificial neural networks. Human
  Brain Mapping  \textbf{40}(17),  4952--4964 (Dec 2019).
  \doi{10.1002/hbm.24750}

\bibitem{jonsson_brain_2019}
Jonsson, B.A., Bjornsdottir, G., Thorgeirsson, T.E., Ellingsen, L.M., Walters,
  G.B., Gudbjartsson, D.F., Stefansson, H., Stefansson, K., Ulfarsson, M.O.:
  Brain age prediction using deep learning uncovers associated sequence
  variants. Nature Communications  \textbf{10}(1), ~5409 (Dec 2019).
  \doi{10.1038/s41467-019-13163-9}

\bibitem{kondo_age_2015}
Kondo, C., Ito, K., Wu, K., Sato, K., Taki, Y., Fukuda, H., Aoki, T.: An age
  estimation method using brain local features for {T1}-weighted images. In:
  2015 37th {Annual} {International} {Conference} of the {IEEE} {Engineering}
  in {Medicine} and {Biology} {Society} ({EMBC}). pp. 666--669. IEEE, Milan
  (Aug 2015). \doi{10.1109/EMBC.2015.7318450}

\bibitem{lam_3d_2020}
Lam, P., Zhu, A.H., Gari, I.B., Jahanshad, N., Thompson, P.M.: 3d
  grid-attention networks for interpretable age and alzheimer's disease
  prediction from structural mri (2020)

\bibitem{lee_deep_2022}
Lee, J., Burkett, B.J., Min, H.K., Senjem, M.L., Lundt, E.S., Botha, H.,
  Graff-Radford, J., Barnard, L.R., Gunter, J.L., Schwarz, C.G., et~al.: Deep
  learning-based brain age prediction in normal aging and dementia. Nature
  Aging  \textbf{2}(5),  412--424 (May 2022). \doi{10.1038/s43587-022-00219-7},
  number: 5 Publisher: Nature Publishing Group

\bibitem{liem_predicting_2017}
Liem, F., Varoquaux, G., Kynast, J., Beyer, F., Kharabian~Masouleh, S.,
  Huntenburg, J.M., Lampe, L., Rahim, M., Abraham, A., Craddock, R.C., et~al.:
  Predicting brain-age from multimodal imaging data captures cognitive
  impairment. NeuroImage  \textbf{148},  179--188 (Mar 2017).
  \doi{10.1016/j.neuroimage.2016.11.005}

\bibitem{ly_improving_2020}
Ly, M., Yu, G.Z., Karim, H.T., Muppidi, N.R., Mizuno, A., Klunk, W.E.,
  Aizenstein, H.J.: Improving brain age prediction models: incorporation of
  amyloid status in alzheimer's disease. Neurobiology of Aging  \textbf{87},
  44--48 (2020). \doi{10.1016/j.neurobiolaging.2019.11.005}

\bibitem{mann_whitney_1947}
Mann, H.B., Whitney, D.R.: {On a Test of Whether one of Two Random Variables is
  Stochastically Larger than the Other}. The Annals of Mathematical Statistics
  \textbf{18}(1),  50 -- 60 (1947). \doi{10.1214/aoms/1177730491}

\bibitem{marcel_torchvision_2010}
Marcel, S., Rodriguez, Y.: Torchvision the machine-vision package of torch. In:
  Proceedings of the 18th {ACM} international conference on {Multimedia}. pp.
  1485--1488. ACM, Firenze Italy (Oct 2010). \doi{10.1145/1873951.1874254}

\bibitem{menze_multimodal_2015}
Menze, B.H., Jakab, A., Bauer, S., Kalpathy-Cramer, J., Farahani, K., Kirby,
  J., Burren, Y., Porz, N., Slotboom, J., Wiest, R., et~al.: The {Multimodal}
  {Brain} {Tumor} {Image} {Segmentation} {Benchmark} ({BRATS}). IEEE
  Transactions on Medical Imaging  \textbf{34}(10),  1993--2024 (Oct 2015).
  \doi{10.1109/TMI.2014.2377694}

\bibitem{more_brain_2023}
More, S., Antonopoulos, G., Hoffstaedter, F., Caspers, J., Eickhoff, S.B.,
  Patil, K.R.: Brain-age prediction: A systematic comparison of machine
  learning workflows. NeuroImage  \textbf{270},  119947 (2023).
  \doi{10.1016/j.neuroimage.2023.119947}

\bibitem{peng_accurate_2021}
Peng, H., Gong, W., Beckmann, C.F., Vedaldi, A., Smith, S.M.: Accurate brain
  age prediction with lightweight deep neural networks. Medical Image Analysis
  \textbf{68},  101871 (Feb 2021). \doi{10.1016/j.media.2020.101871}

\bibitem{petersen_alzheimers_2010}
Petersen, R.C., Aisen, P.S., Beckett, L.A., Donohue, M.C., Gamst, A.C., Harvey,
  D.J., Jack, C.R., Jagust, W.J., Shaw, L.M., Toga, A.W., et~al.: Alzheimer's
  {Disease} {Neuroimaging} {Initiative} ({ADNI}): {Clinical} characterization.
  Neurology  \textbf{74}(3),  201--209 (Jan 2010).
  \doi{10.1212/WNL.0b013e3181cb3e25}

\bibitem{poloni_deep_2022}
Poloni, K.M., Ferrari, R.J.: A deep ensemble hippocampal {CNN} model for brain
  age estimation applied to {Alzheimer}’s diagnosis. Expert Systems with
  Applications  \textbf{195},  116622 (Jun 2022).
  \doi{10.1016/j.eswa.2022.116622}

\bibitem{popescu_local_2021}
Popescu, S.G., Glocker, B., Sharp, D.J., Cole, J.H.: Local {Brain}-{Age}: {A}
  {U}-{Net} {Model}. Frontiers in Aging Neuroscience  \textbf{13},  761954 (Dec
  2021). \doi{10.3389/fnagi.2021.761954}

\bibitem{raghu_transfusion_2019}
Raghu, M., Zhang, C., Kleinberg, J., Bengio, S.: Transfusion: Understanding
  Transfer Learning for Medical Imaging. Curran Associates Inc., Red Hook, NY,
  USA (2019), \url{https://dl.acm.org/doi/10.5555/3454287.3454588}

\bibitem{navab_u-net_2015}
Ronneberger, O., Fischer, P., Brox, T.: U-{Net}: {Convolutional} {Networks} for
  {Biomedical} {Image} {Segmentation}. In: Navab, N., Hornegger, J., Wells,
  W.M., Frangi, A.F. (eds.) Medical {Image} {Computing} and
  {Computer}-{Assisted} {Intervention} – {MICCAI} 2015, vol.~9351, pp.
  234--241. Springer International Publishing, Cham (2015).
  \doi{10.1007/978-3-319-24574-4\_28}, series Title: Lecture Notes in Computer
  Science

\bibitem{russakovsky_imagenet_2015}
Russakovsky, O., Deng, J., Su, H., Krause, J., Satheesh, S., Ma, S., Huang, Z.,
  Karpathy, A., Khosla, A., Bernstein, M., et~al.: {ImageNet} {Large} {Scale}
  {Visual} {Recognition} {Challenge}. International Journal of Computer Vision
  \textbf{115}(3),  211--252 (Dec 2015). \doi{10.1007/s11263-015-0816-y}

\bibitem{wang_age_2014}
Wang, J., Li, W., Miao, W., Dai, D., Hua, J., He, H.: Age estimation using
  cortical surface pattern combining thickness with curvatures. Medical \&
  Biological Engineering \& Computing  \textbf{52}(4),  331--341 (Apr 2014).
  \doi{10.1007/s11517-013-1131-9}

\bibitem{wyman_standardization_2013}
Wyman, B.T., Harvey, D.J., Crawford, K., Bernstein, M.A., Carmichael, O., Cole,
  P.E., Crane, P.K., DeCarli, C., Fox, N.C., Gunter, J.L., et~al.:
  Standardization of analysis sets for reporting results from {ADNI} {MRI}
  data. Alzheimer's \& Dementia  \textbf{9}(3),  332--337 (May 2013).
  \doi{10.1016/j.jalz.2012.06.004}

\bibitem{zhang_road_2018}
Zhang, Z., Liu, Q., Wang, Y.: Road extraction by deep residual u-net. IEEE
  Geoscience and Remote Sensing Letters  \textbf{15}(5),  749--753 (2018).
  \doi{10.1109/LGRS.2018.2802944}

\end{thebibliography}






\end{document}